\documentclass[sigconf]{acmart}
\usepackage{multirow}
\usepackage{float}
\usepackage{graphicx}
\usepackage{amsmath}
\usepackage{algorithmic}

\newcommand{\vect}[1]{\mathbf{#1}}   
\newcommand{\mat}[1]{\mathbf{#1}}    
\newcommand{\appropto}{\mathrel{\vcenter{
  \offinterlineskip\halign{\hfil$##$\cr
    \propto\cr\noalign{\kern2pt}\sim\cr\noalign{\kern-2pt}}}}}

\AtBeginDocument{%
  \providecommand\BibTeX{{%
    \normalfont B\kern-0.5em{\scshape i\kern-0.25em b}\kern-0.8em\TeX}}}

\copyrightyear{2021}
\acmYear{2021}
\setcopyright{acmcopyright}\acmConference[MM '21]{Proceedings of the 29th ACM International Conference on Multimedia}{October 20--24, 2021}{Virtual Event, China}
\acmBooktitle{Proceedings of the 29th ACM International Conference on Multimedia (MM '21), October 20--24, 2021, Virtual Event, China}
\acmPrice{15.00}
\acmDOI{10.1145/3474085.3481540}
\acmISBN{978-1-4503-8651-7/21/10}

\begin{document}
\fancyhead{}
\title[Unifying Multimodal Transformer for Bi-directional Image and Text Generation]{Unifying Multimodal Transformer \\ for Bi-directional Image and Text Generation}

\author{Yupan Huang}
\authornote{Work done during an internship at Microsoft Research Asia.}
\affiliation{
\institution{Sun Yat-sen University}
\country{}
}
\email{huangyp28@mail2.sysu.edu.cn}

\author{Hongwei Xue}
\authornotemark[1]
\affiliation{\institution{University of Science and Technology of China}
\country{}}
\email{gh051120@mail.ustc.edu.cn}

\author{Bei Liu}
\affiliation{\institution{Microsoft Research Asia}
\country{}}
\email{Bei.Liu@microsoft.com}

\author{Yutong Lu}
\authornote{Corresponding Author.
}
\affiliation{\institution{Sun Yat-sen University}
\country{}}
\email{luyutong@mail.sysu.edu.cn}

\renewcommand{\shortauthors}{Huang, et al.}

\begin{abstract}
We study the joint learning of image-to-text and text-to-image generations, which are naturally bi-directional tasks.
Typical existing works design two separate task-specific models for each task, which impose expensive design efforts.
In this work, we propose a unified image-and-text generative framework based on a single multimodal model to jointly study the bi-directional tasks.
We adopt Transformer as our unified architecture for its strong performance and task-agnostic design.
Specifically, we formulate both tasks as sequence generation tasks, where we represent images and text as unified sequences of tokens, and the Transformer learns multimodal interactions to generate sequences.
We further propose two-level granularity feature representations and sequence-level training to improve the Transformer-based unified framework.
Experiments show that our approach significantly improves previous Transformer-based model X-LXMERT's FID from 37.0 to 29.9 (lower is better) for text-to-image generation, and improves CIDEr-D score from 100.9\% to 122.6\% for fine-tuned image-to-text generation on the MS-COCO dataset.
Our code is available online.

\end{abstract}



\maketitle

\begin{figure}[ht]
\small
    \centering
    \includegraphics[width=0.99\linewidth]{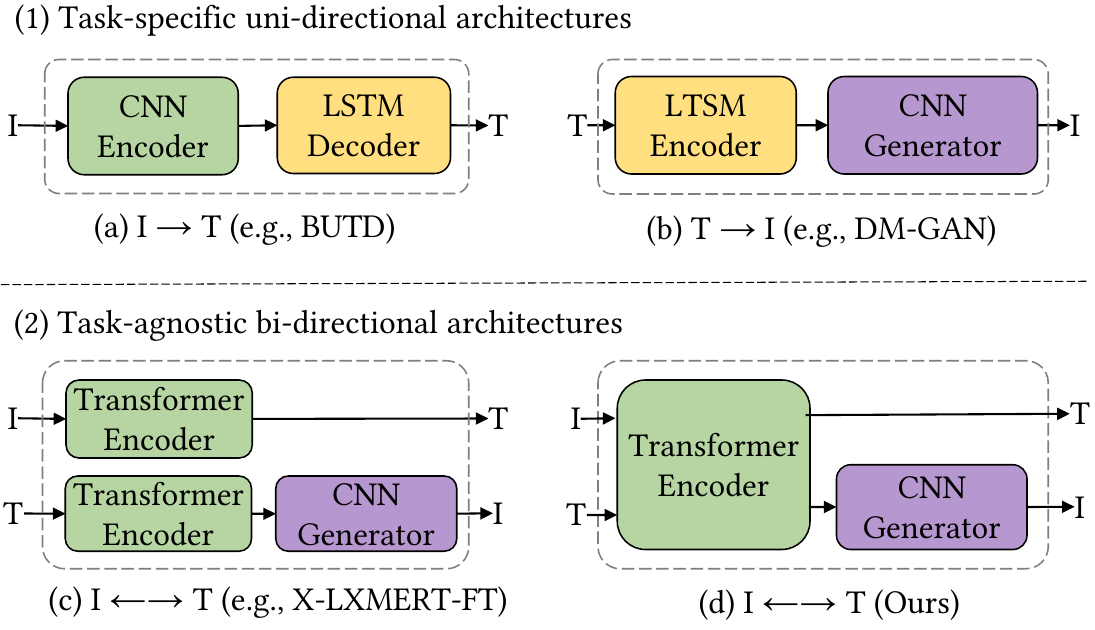}
    \caption{Comparison with existing works~\cite{anderson2018bottom,zhu2019dm,cho2020x} on bi-directional image and text generation tasks. 
    Our task-agnostic bi-directional architecture as show in (2) releases design efforts of task-specific architectures in (1).
    When comparing with other Transformer-based models in a pre-train and fine-tune fashion as shown in (c), our model unifies two tasks in a Transformer model.
    ``X-LXMERT-FT'' denotes fine-tuning X-LXEMRT~\cite{cho2020x} for two tasks. ``I'' and ``T'' denote ``image'' and ``text'' respectively.} \label{fig:intro}
\end{figure}

\section{Introduction}
As a result of developing multimodal processing technology, networks, and devices, the number of existing multimodal industrial applications (especially mobile applications) is rapidly increasing.
This trend has encouraged the development of techniques to support multimodal interaction in a unified system.
Bi-directional image and text generation is a technique that automatically translates across real scene images and natural language descriptions.
It has a broad range of industrial applications.
For example, a product description-and-picture generation system is an important application for consumers to search and preview products.
Specifically, consumers could easily get a text query when the system receives a picture of one or several products.
Meanwhile, the users could say ``show me the picture of a room with wooden furniture and purple curtains.'' and receive an illustrating picture by the system to preview this collocation.
As these applications are equipped in mobile devices, unifying multimodal interactions in a single model
will be a better choice to optimize storage utilization compared to two separate models.

Despite the great benefit of a unified framework for image and text generation, the bi-directional generation tasks are conducted separately with task-specific architectures traditionally.
As shown in Figure~\ref{fig:intro} (1), a typical image-to-text generator consists of a visual encoder (e.g., CNN) to embed visual information, and a caption decoder (e.g., LSTM) to generate captions~\cite{anderson2018bottom};
while predominant text-to-image generators adopt Generative Adversarial Nets (GANs) \cite{NIPS2014_5ca3e9b1} framework based on CNN architectures~\cite{zhu2019dm}.
To support multimodal interaction, Huang ~\textit{et al.} jointly trains an LSTM-based image-to-text generator and a GAN-based text-to-image generator in a framework~\cite{huang2018turbo}.
However, task-specific architectures are still needed, which introduces expensive design efforts.

To alleviate above hassles, in this paper, we propose to unify image-to-text and text-to-image generation tasks in one framework.
In this framework, we adopt Transformer-based architecture since it supports simple and task-agnostic designs, and exhibits strong performance in image or text generative models~\cite{li2020oscar,ramesh2021zero}.
We formulate both tasks as sequence generation tasks, where an image and a text are represented as sequences of tokens, and the model learns to predict target tokens conditioned on other ground-truth tokens with a cross-entropy loss training.
Existing Transformer-based text-to-image generation works~\cite{cho2020x,ramesh2021zero,Ding2021CogViewMT} can be extended to support image-to-text generation by exchanging the order of text and image tokens in their input sequences.
While they have shown some initial promise, these approaches still exhibit two major challenges for bi-directional generations: information loss caused by the feature discretization process, and error accumulation caused by the cross-entropy loss training.
Specifically, first, Transformer-based approaches enable image generation by clustering dense image features into discrete indices as the labels of image tokens~\cite{cho2020x,ramesh2021zero,Ding2021CogViewMT}, However, this discretization process is harmful to image-to-text generation due to information loss.
Second, the cross-entropy loss training in a ``teacher-forcing'' manner creates a mismatch between training and testing, as the model is only exposed to the training data distribution instead of its own prediction.
This ``exposure bias'' results in error accumulation at test time~\cite{ranzato2015sequence}.
Due to these challenges, the typical Transformer-based approach, X-LXMERT~\cite{cho2020x}, generates captions weakly correlated with their source images, and even its text-to-image generation results on automatic metric are worse than its comparing GAN-based approach~\cite{zhu2019dm}.

We address these challenges with two major designs, i.e., two-level granularity feature representations and sequence-level training.
First, we introduce two-level granularity feature representations, in which we use dense features to reduce information loss for image-to-text generation, and discrete features to enable text-to-image generation.
Second, we propose a training strategy that optimizes our model based on its sequence-level prediction instead of token-level predictions to bridge the gap between training and testing.
Based on this strategy, we particularly introduce a CLIP-based image-level loss for text-to-image generation, which improves the consistency between generated images and the source text by leveraging a large-scale pre-trained multimodal model CLIP~\cite{radford2021learning}.
Moreover, because CLIP learns from a vast amount of image-text pairs on the internet and releases task-specific crowd-sourced labeling, we naturally propose a CLIP-based metric for text-to-image evaluation.

We share most of the Transformer networks for image and text generation tasks and train them iteratively as depicted in Figure~\ref{fig:intro} (2).
This paradigm facilitates image-and-text shared embedding learning, which improves the performance with a half of model size when compared with two separate Transformer models.
Compared with a previous Transformer-based approach, X-LXMERT, our approach significantly improves CLIPScore from 72.9 to 77.2 for text-to-image generation, and improves CIDEr-D score from 100.9\% to 122.6\% for image-to-text generation on the MS-COCO dataset.
In summary, our contributions are three-fold:
\begin{itemize}
    \item We present a unified image-and-text generative framework based on a Transformer model with two proposals: (1) two-level granularity feature representation to avoid information loss
    ; (2) sequence-level training to mitigate the gap between training and testing.
    \item We leverage the powerful pre-trained model CLIP to improve text-to-image generation consistency, and to facilitate its evaluation without extra crowd-sourced labeling.
    \item We conduct automatic and human evaluations that demonstrates our approach significantly improves the quality of both tasks over prior approaches on the MS-COCO dataset.
\end{itemize}

\begin{figure*}[t]
    \centering
    \includegraphics[width=0.99\linewidth]{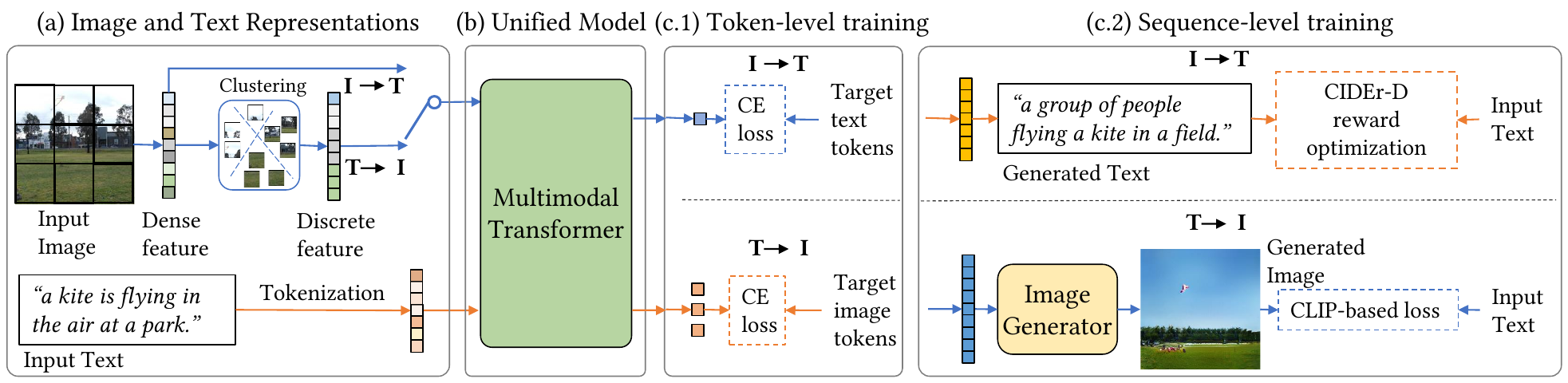}
    \caption{Overview of our framework. We formulate image and text generation tasks as sequence generation tasks. (a) Image-and-text pairs are represented as unified token sequences. Specifically, images are represented as two-level granularity features where we use dense features to reduce information loss for I->T and discrete features to enable image tokenization for T->I.
    (b) The unified Transformer learns multimodal interactions to generate sequences.
    (c) In addition to the standard token-level training, we introduce sequence-level training strategy to mitigate the mismatch between training and testing.
    ``I'' and ``T'' denote ``image'' and ``text''.} \label{fig:framework}
\end{figure*}

\section{Related Work}

\subsection{Image-to-Text Generation}
The task of image captioning has attracted increasing attention\cite{rennie2017self,anderson2018bottom,lu2018neural,yao2018exploring,yang2019learning,li2019entangled,huang2019attention,guo2019aligning,yang2019auto,cornia2019m,zhou2020unified,li2020oscar}.
The dominant image captioning models adopt encoder-decoder architectures, which embed images with a CNN encoder, and generate captions with an RNN~\cite{rennie2017self,anderson2018bottom} or Transformer decoder~\cite{vaswani2017attention,cornia2019m}.
By connecting the encoder and decoder with attention mechanisms, the decoder can selectively integrate relevant information for word generation.
Transformer has shown up prominently thanks to its capability in parallel training and modeling global dependency relying entirely on an attention mechanism~\cite{vaswani2017attention}.
Transformer-based large-scale vision-and-language pre-training works over numerous image-text pairs have shown great advances for image captioning~\cite{zhou2020unified,li2020oscar}.

\subsection{Text-to-Image Generation}
Synthesizing images from text descriptions continues to be challenging.
Deep generative models based on pixelCNN~\cite{van2016conditional}, approximate Langevin sampling~\cite{nguyen2017plug}, variational autoencoders~\cite{kingma2014auto}, and Generative Adversarial Nets (GANs)~\cite{NIPS2014_5ca3e9b1} have been proposed.
Dominant works adopt the GAN framework, which usually consists of a generator and a discriminator to solve a min-max optimization problem, and has shown effectiveness in image generations~\cite{reed2016generative,zhang2017stackgan,xu2018attngan,zhu2019dm}.
Typical works generate images with multiple stages, that they firstly sample low-resolution images, then they gradually upsample and improve images in later stages~\cite{reed2016generative,zhang2017stackgan,xu2018attngan,zhu2019dm}. 
Recent large-scale Transformer-based pre-trained works impressed us with their high fidelity images~\cite{ramesh2021zero,Ding2021CogViewMT}.
DALL-E and CogView greatly advance the quality of text-to-image generation by training transformers with 12-billion and 4-billion parameters on 250 million and 30 million image-text pairs respectively.

\subsection{Bi-directional Image-and-Text Generation}
Image-to-text and text-to-image generation are bi-directional tasks.
Huang \textit{et al.}~\cite{huang2018turbo} propose to jointly train an image-to-text generator based on an LSTM network and a text-to-image generator based on a GAN network to boost both generators by enforcing cycle-consistency in turbo learning.
MirrorGAN~\cite{qiao2019mirrorgan} focus on the text-to-image generation task and uses an off-the-shelf captioning model to regularize the redescription of the generated images to be semantic consistent with the original text.
X-LXMERT~\cite{cho2020x} is a pre-trained model based on Transformer for text-to-image generation, visual question answering, and visual reasoning.
Different from these works that use separate bi-directional models, in this paper, we use the same model for image and text generation tasks.

\section{Model}
In this section, we introduce our unified multimodal framework and the design of two-level granularity image representation.
Figure~\ref{fig:framework} (a, b) gives an overview of the model.

\subsection{Unified Multimodal Framework}
Our model mainly consists of a unified multimodal Transformer~\cite{vaswani2017attention}, which has a multi-layer architecture and each layer mainly consists of multi-head self-attention and position-wise fully connected feed-forward networks.
We adopt LXMERT~\cite{tan2019lxmert} as our Transformer-based architecture following X-LXMERT~\cite{cho2020x} for a direct comparison, since our proposal is mainly based on feature representation and training mechanism but not on the specific model.
LXMERT is a cross-modality Transformer consisting of an object-relation encoder, a language encoder and a cross-modality encoder.
We omit an exhaustive background description of the model architecture and refer readers to \cite{vaswani2017attention,tan2019lxmert,cho2020x} for additional details.

To enable both image-to-text and text-to-image generation tasks, we formulate both tasks as sequence generation tasks.
Specifically, we firstly pre-process image-and-text pairs into sequences of image tokens and text tokens.
Then the Transformer accepts sequences with masked tokens as input, and maps the input to a contextualized representation via attention networks.
Lastly, an image or text linear transformation classifier projects the contextualized representation to predicted target tokens.
In our model, we share the parameters of most Transformer modules, except for the last classifiers since image and text vocabularies have different sizes and semantics.
For text-to-image generation, we use an additional GAN-based image generator to convert the size of $8\times 8$ image token predictions to a $256 \times 256$ resolution visible image.

{Text representation} is a combination of the position embedding and the word embedding, where the position refers to the index of the word within the caption, and the word embedding is initialized from pre-trained models like BERT~\cite{devlin2019bert} or LXMERT~\cite{tan2019lxmert}.

\subsection{Two-level Granularity Image Representations}\label{subsection:image_feature}
{Image representation} is a combination of image features and position features.
\textit{For image features,} we split an image into a sequence of uniform grid-level patches, which are effective to learn visual representations for vision-and-language tasks~\cite{jiang2020defense,cho2020x,huang2020pixel,huang2021seeing}. 
We propose to use two-level granularity image features.
\textbf{(1) Fine-grained dense features:}
We extract the grid features with a Faster R-CNN~\cite{ren2015faster} object detector pre-trained on the Visual Genome dataset~\cite{krishna2017visual}. 
The dense grid features are used as visual inputs for image-to-text generation to reduce the loss of image information.
\textbf{(2) Coarse-grained discrete features:}
We use discrete clustering features of the dense features to construct the ground-truth labels of visual tokens following X-LXMERT~\cite{cho2020x}.
The discretization process is critical in Transformer-based image generation methods since it is hard to construct a large visual vocabulary for the diverse image pixels or image features, while discretization helps reduce the feature noise and vocabulary size ~\cite{cho2020x,ramesh2021zero}.
Specifically, we firstly create a visual vocabulary using K-mean clustering, approximate the target visual features via a nearest neighbor search, and then obtain the cluster indices and discrete grid features for each image.
Naturally, \textit{the position feature} of each grid is a 2-D embedding of the grid-level bounding box positions.

\section{Training}
We formulate the bi-directional image and text generation tasks as sequence generation tasks and iteratively train the Transformer.
Sequence models are usually trained using the cross-entropy loss in a ``Teacher-Forcing'' manner, where ground truth tokens are given in each step of training.
Since the model is only exposed to the training data distribution instead of its own predictions, a mismatch between training and testing named ``exposure bias'' is created and results in error accumulation at test time~\cite{ranzato2015sequence}.
To tackle this problem, we design a two-stage training strategy.
\begin{itemize}
    \item \textbf{Stage 1. Token-level Training.} This stage aims to generate fluent captions and realistic images by ``teacher-forcing'' training on \textbf{word level} or \textbf{grid level} in each step for image-to-text or text-to-image generation respectively.
    \item \textbf{Stage 2. Sequence-level Training.} This stage further optimizes the model with generated \textbf{text} or \textbf{image} sequence to bridge the gap between training and testing.
\end{itemize}

\noindent Figure~\ref{fig:framework} (c) gives an overview of the training process.

\subsection{Image-to-Text Generation}

\subsubsection{Stage 1. Word-level Training}
In the first stage, we aim to train an image caption generator in a "Teacher-Forcing" manner similar to training a uni-directional autoregressive transformer.
The model is trained to maximize the likelihood of the next ground-truth word given the previous ground-truth words and visual context using back-propagation.
We denote parameters of Transformer model with $\theta$ and minimize the following cross-entropy loss:
\begin{align}
L^T_{Tok}(\theta) &= -\sum_{t=1}^{L} \log p_{\theta}(\vect{y}_l \mid \vect{y}_{1:l-1}, \mat{X}),
\end{align}
\noindent where $\mat{X}=\vect{x}_{1:M}$ and $\mat{Y}=\vect{y}_{1:L}$ are the sequences of ground truth image and text tokens respectively. $M$ and $L$ are the sequence length for image and text sequences respectively.

\subsubsection{Stage 2. Sentence-level Training}
To alleviate the exposure bias, we adopt the reinforcement learning algorithm of Self-Critical Sequence Training (SCST)~\cite{rennie2017self} to directly optimize non-differentiable sentence-level metrics (e.g., CIDEr-D~\cite{vedantam2015cider}).
We minimize the negative expected score function $r$ of CIDEr-D metric:
\begin{align}
\label{eqn:ciderloss}
L^{T}_{Seq}(\theta) &= -\ensuremath\textbf{E}_{\vect{\hat{y}}_{1:L} \sim p_\theta}[r(\vect{\hat{y}}_{1:L})],
\end{align}
where $\mat{\hat{Y}}=\vect{\hat{y}}_{1:L}$ denotes the sampled text tokens.

\subsection{Text-to-Image Generation}

\subsubsection{Stage 1. Grid-level Training.}
The training target is to maximize the likelihood of the target visual tokens given the other ground-truth visual tokens and text context, which is similar to the training target of the first stage of image-to-text generation.
However, text-to-image direction aims to predict multiple tokens instead of one token at each step, by minimizing the cross-entropy loss overall $M'$ masked tokens:
\begin{align}
L^I_{Tok}(\theta) &= -\sum_{m=1}^{M'}\log p_{\theta}(\vect{x}_m \mid \vect{x}_{\backslash M'}, \mat{Y}),
\end{align}
where $\vect{x}_{\backslash M'}$ denotes the sequence of tokens excluding the masked tokens.
We follow X-LXMERT to use this training strategy to enables a  non-autoregressive sampling strategy, i.e., mask-predict-k strategy~\cite{ghazvininejad2019mask}.
In this way, only a few sampling steps (e.g., $k=4$) are needed to generate all visual tokens of an image, which enjoys a faster inference speed for broader industrial applications.

\begin{table*}[]
\centering
\caption{Ablations on image-to-text (I2T) and text-to-image (T2I) tasks on MSCOCO test set.
(1) Our unified single Transformer model exhibits comparable performance to task-specific Transformers with half of model parameters.
(2) Our two-level granularity image representation inherits the advantages from both discrete features for T2I, and dense features for I2T. 
(3) Our proposed sequence-level training significantly improves both tasks.
(4) Our proposed CLIP-based loss improves T2I by improving the consistency between generated images and the source text.
}
\label{tab:ablation_fullmodel}
\begin{tabular}{lcccccccccc}
\toprule
\multirow{3}{*}{Model} & \multirow{3}{*}{Parameters$\downarrow$} & \multicolumn{5}{c}{Image-to-Text Generation} & \multicolumn{4}{c}{Text-to-Image Generation} \\
 &  & \multirow{2}{*}{B@4$\uparrow$} & \multirow{2}{*}{M$\uparrow$} & \multirow{2}{*}{R$\uparrow$} & \multirow{2}{*}{C$\uparrow$} & \multirow{2}{*}{S$\uparrow$} & \multirow{2}{*}{CLIPScore$\uparrow$} & \multirow{2}{*}{FID$\downarrow$} & \multicolumn{2}{c}{R-precision (hard/easy)} \\
 &  &  &  &  &  &  &  &  & ViLBERT$\uparrow$ & CLIP$\uparrow$ \\ \hline
\textbf{Ours} & \textbf{228M} & 37.3 & 28.2 & 57.9 & \textbf{122.6} & 21.9 & \textbf{77.2} & \textbf{29.9} & \textbf{37.7}/59.2 & {\textbf{40.7/69.9}} \\
w/o unified architecture & 456M & 37.4 & 28.2 & 58.0 & 122.3 & 22.0 & {76.5} & {30.2} & 37.0/\textbf{59.6} & 40.3/68.9 \\
w/o two-level features & \multicolumn{1}{l}{} & \multicolumn{1}{l}{} & \multicolumn{1}{l}{} & \multicolumn{1}{l}{} & \multicolumn{1}{l}{} & \multicolumn{1}{l}{} & \multicolumn{1}{l}{} & \multicolumn{1}{l}{} & \multicolumn{1}{l}{} & \multicolumn{1}{l}{} \\
\multicolumn{1}{c}{\textit{Dense feature}} & 228M & 37.2 & 28.2 & 57.9 & 122.2 & \textbf{22.0} & 75.7 & 34.9 & 32.9/51.2 & 38.6/61.9 \\
\multicolumn{1}{c}{\textit{Discrete feature}} & 228M & 34.7 & 27.0 & 56.0 & 114.3 & 20.7 & 76.9 & 30.2 & 37.0/59.1 & 40.7/{69.3} \\
w/o sequence-level training & 228M & 32.2 & 26.9 & 54.8 & 107.9 & 20.2 & 73.4 & 33.5 & 33.3/51.1 & 35.5/63.0 \\
w/o CLIP-based loss & 228M & \textbf{37.6} & \textbf{28.3} & \textbf{58.1} & 122.5 & 22.0 & 72.5 & 40.1 & 30.7/47.6 & 34.2/59.0 \\
\bottomrule
\end{tabular}
\end{table*}

\subsubsection{Stage 2. Image-level Training.}\label{stage2_image}
Although the cross-entropy loss in grid-level training has shown initial promise, there remain two major issues.
First, the loss imposes restrictive supervision on each generated image to regard one reference image as a ``gold label''.
This violates the one-to-many property of text-to-image mapping, where a caption can correspond to many feasible images.
Second, the loss is based on image grid indices or features, which ignores the relations across grids.

To tackle these issues, we propose to directly optimize the model towards generating an image that is more semantic consistent with the source caption instead of one reference image.
To achieve this, we leverage a large-scale pre-trained multimodal model CLIP~\cite{radford2021learning} to score the image-text consistency.
This is desirable since CLIP is a general model pre-trained on 400M image-text pairs from the web and exhibits unparalleled zero-shot transferable ability in a great variety of classification benchmarks.
DALLE has used CLIP to re-rank its generated images with CLIP-based scores as an offline post-process~\cite{ramesh2021zero}, while we are the first to leverage a CLIP-based loss to directly regularize the training for text-to-image generation.

Specifically, we extract image and text embedding from the CLIP and calculate their cosine similarity to obtain CLIP-based loss:
\begin{equation}
{L}^{I}_{clip}(\theta) =-\mathrm{max}(cos(\mathcal{I}(\mat{\hat{X}}),\mathcal{T}(\mat{Y})),0),
\end{equation}
where $\mathcal{I}(\cdot)$ and $\mathcal{T}(\cdot)$ are the image and text embedding extraction networks of CLIP.
Note that the image embedding is also a Gumbel-Softmax approximation~\cite{jang2016categorical} to support model optimization with back-propagation.
Learning from CLIP has several potential strengths.
It releases the crowd-sourced labeling, and naturally conforms to the one-to-many property of text-to-image mapping thanks to learning from the vast amount of image-text pairs on the internet.
Moreover, it connects text with the whole images instead of image grids to consider the relations across grids, thus it encourages higher semantic consistency between images and text.

We also have experimented with a grid feature similarity loss, a pixel-wise loss and a perceptual loss \cite{johnson2016perceptual}, but we do not observe much improvement in our preliminary experiments.
Due to the high variance property of sampling results of non-autoregressive sampling strategy, we conduct this second stage training accompanying with the first stage training iteratively, since the first stage of ``teacher-forcing'' training can encourage the sampling coherency.

\section{Experiments}
\subsection{Experimental Setup}

\noindent \textbf{MS-COCO Dataset~\cite{lin2014microsoft}.}
We evaluate our proposed method on the popular MS-COCO dataset.
It is collected using Amazon Mechanical Turk (AMT) with five annotated captions for each image.
MS-COCO dataset's official splits include 82,783/40,504/40,775 images for training/validation/testing set respectively.
We train our model on the train split and evaluate our model on the validation split by randomly sampling 30,000 images following most text-to-image generation works~\cite{cho2020x,zhu2019dm}.
For image-to-text generation (image captioning), we follow most captioning works to evaluate our model on the Karpathy test split, which is a subset of the validation set consisting of 5,000 images.
Our results on image captioning are not directly comparable to other image captioning models since they are trained with a larger split (113,287 vs. 82,783) and are expected to score higher.
Moreover, we use grid-based 8x8 features for a fair comparison with X-LXMERT, while this feature is weaker than the 100 region-based feature used by standard image captioning works.

\noindent \textbf{Implementation Details.}
Our code is available online\footnote{\url{https://github.com/researchmm/generate-it}}.
For details on \textbf{model architecture}, we initialize our model from the pre-trained X-LXMERT model~\cite{cho2020x} for a direct comparison, which adopts the architecture of LXMERT~\cite{tan2019lxmert} and is pre-trained with MS-COCO Captions~\cite{lin2014microsoft}, Visual Genome~\cite{krishna2017visual} and VQA~\cite{antol2015vqa} datasets.
We adopt an image generator consisting of convolutional layers and trained with Generative Adversarial Networks (GAN) \cite{NIPS2014_5ca3e9b1} method following X-LXMERT.
Also, we directly use the pre-trained image generator provided by X-LXMERT for a fair comparison\footnote{\url{https://github.com/allenai/x-lxmert}}.
We limit the length of a text caption to $L=17$ tokens, and the grid size of each image is $M=8\times 8$.
We use a vocabulary of 30,522 tokens for text words, and a vocabulary of 10,000 tokens for image grids.

For details on \textbf{training}, 
we use the AdamW~\cite{kingma2014adam} optimizer with
beta coefficients of 0.9 and 0.999,
and a weight decay of 1e-2 following X-LXMERT.
Both image-to-text or text-to-image generation tasks take 100,000 iterations in the first or second stage training.
For the first stage training, we linearly warm up the learning rate from 0 to 5e-5 over the first 5\% iterations, and cosine decay it in the rest of training steps following X-LXMERT.
Since the second stage of training is initialized from the first stage, we use a fixed smaller learning rate of 1e-6.
We train the first stage with a batch size of 256, and the second stage of 160 empirically.
We use a label smoothing of 0.2~\cite{szegedy2016rethinking}, and a gradient clipping threshold of 1.0.
We adopt mixed-precision training to reduce memory cost and speed up the training procedure.

\begin{table*}[]
\centering
\caption{Comparisons with existing approaches on MS-COCO test set.
We generate images (or captions) for DM-GAN~\cite{zhu2019dm}, X-LXMERT~\cite{cho2020x} with their released codes and models, and evaluate the images (or captions) in the same way as ours and X-LXMERT-FT for more direct comparison.
For BUTD~\cite{anderson2018bottom} and Turbo-RL~\cite{huang2018turbo}, we report the numbers recorded in their published papers.
``-'' indicates the detail is not reported.
``N/A'' is the abbreviation of ``not applicable''.
``I'', ``T'', ``Arch'' and ``TF'' denote ``image'', ``text'', ``architecture'' and ``Transformer'' respectively.
}
\label{tab:sota}
\begin{tabular}{ccccccccccccc}
\toprule

\multirow{3}{*}{Direction} & \multirow{3}{*}{Model} & \multicolumn{6}{c}{Image-to-Text Generation} & \multicolumn{5}{c}{Text-to-Image Generation} \\
 &  & \multirow{2}{*}{Arch} & \multirow{2}{*}{B@4$\uparrow$} & \multirow{2}{*}{M$\uparrow$} & \multirow{2}{*}{R$\uparrow$} & \multirow{2}{*}{C$\uparrow$} & \multirow{2}{*}{S$\uparrow$} & \multirow{2}{*}{Arch} & \multirow{2}{*}{CLIPScore$\uparrow$} & \multirow{2}{*}{FID$\downarrow$} & \multicolumn{2}{c}{R-precision (hard/easy)} \\
 &  &  &  &  &  &  &  &  &  &  & ViLBERT$\uparrow$ & CLIP$\uparrow$ \\ \hline
T-\textgreater{}I & DM-GAN & \multicolumn{6}{c}{N/A} & CNN & 72.3 & 35.9 & 32.7/58.4 & 34.4/69.1 \\
I-\textgreater{}T & BUTD$^\dagger$ & RNN & 36.3 & 27.7 & 56.9 & 120.1 & 21.4 & \multicolumn{5}{c}{N/A} \\ \hline
\multirow{4}{*}{I\textless{}--\textgreater{}T} & Turbo-RL$^\dagger$ & RNN & 31.6 & 21.9 & 49.8 & 74.8 & 17.5 & CNN & - & - & - & - \\
 & X-LXMERT & TF & 15.2 & 22.7 & 42.2 & 41.0 & 16.6 & TF & 72.9 & 37.0 & 33.1/49.8 & 36.3/60.1 \\
 & X-LXMERT-FT & TF & 30.2 & 25.8 & 53.3 & 100.9 & 19.1 & TF & 73.3 & 33.9 & 33.1/50.7 & 35.1/62.3 \\
 & \textbf{Ours} & TF & \textbf{37.3} & \textbf{28.2} & \textbf{57.9} & \textbf{122.6} & \textbf{21.9} & TF & \textbf{77.2} & \textbf{29.9} & \textbf{37.7/59.2} & \textbf{40.7/69.9} \\ 
\bottomrule
\multicolumn{13}{l}{\footnotesize 
Note that the models denoted with $^\dagger$ are trained on 113,287 images, and are expected to score higher on image-to-text generation than other models trained on 82,783 images.}

\end{tabular}
\end{table*}

\subsection{Evaluation Criteria} 

\noindent \textbf{Image-to-Text Generation.}
We report five commonly-used automatic evaluation metrics for the image captioning task:
BLEU@N~\cite{papineni2002bleu} (N=1,4), ROUGE-L~\cite{lin2004rouge}, MEREOR~\cite{banerjee2005meteor}, CIDEr-D~\cite{vedantam2015cider} and SPICE~\cite{anderson2016spice}, which are denoted by B@N, M, R, C, and S for abbreviation.

\noindent \textbf{Text-to-Image Generation.}
We evaluate text-to-image generation from three aspects:
(1) \textbf{Fréchet Inception Distance (FID)}~\cite{NIPS2017_8a1d6947}
compares the distribution of generated images with the distribution of real images.
A lower FID implies a closer distance between generated image distribution and real-world image distribution.
(2) \textbf{R-precision}~\cite{xu2018attngan}
ranks retrieval results to evaluate whether the generated image is well conditioned on the given text.
We follow X-LXMERT to use two variants of R-precision to fully evaluate our model.
The \textit{easy} variant randomly samples negatives among the test caption set.
The \textit{hard} variant swaps a word in a caption with another word within the same category.
To compute the image and text similarity for ranking, X-LXMERT uses ViLBERT-MT~\cite{lu2019vilbert}, an off-the-shelf multimodal network based on object-level and word-level representations.
We propose to use CLIP-based representations, as a complementary R-precision evaluation metric from global image and text levels.
(3) \textbf{CLIPScore}. 
Since the R-precision cannot directly reflect the individual image-and-text consistency, we propose to use a CLIP-based score, which calculates the cosine similarity between the image and text representations from CLIP~\cite{radford2021learning}, as a complementary metric for text-to-image evaluation.
CLIP is a powerful multimodal pre-trained model to evaluate the image-text consistency, where a CLIP-based metric (i.e., CLIPScore) has been proposed by Jack ~\textit{et al.} to evaluate image captioning models without reference captions~\cite{hessel2021clipscore}.
Instead, we extend CLIPScore to evaluate image generation models.

We do not use Inception score (IS) \cite{NIPS2016_8a3363ab}, since it overfits within the context of text-to-image generation and can be manipulated to achieve much higher scores using simple tricks~\cite{barratt2018note,hinz2020semantic}.

\subsection{Ablation Studies}
We conduct ablation studies to evaluate our designs.
Except for the ablated component, we use the same model, dataset and other implementation details.
Results are shown in Table~\ref{tab:ablation_fullmodel}.

\noindent \textbf{What is the benefit of unified architecture?}
We evaluate the impact of training a bi-directional unified Transformer by comparing our model to two separate Transformer models for image-to-text and text-to-image generations.
For text-to-image generation task, our unified model outperforms the separate model on most metrics.
They exhibit comparable performance on image-to-text generation task.
Moreover, the parameter size of our unified model is a half of two separate models' parameter size.
We expect smaller Transformer could further decrease the model size.
Thus our model significantly benefits industrial application towards optimizing storage utilization.

\noindent \textbf{Do two-level granularity image features help?}
As we have introduced in Section \ref{subsection:image_feature}, the common practice for image-to-text or text-to-image generation is using images' original dense feature or their discrete form respectively.
To conduct bi-directional generations in a unified model, we separately train three models with inputs of dense feature, discrete feature and our two-level granularity feature.
Note that when taking dense feature as input, the model still learns to predict discrete image tokens instead of dense features for text-to-image generation, otherwise the model would fail to generate normal images validated by X-LXMERT~\cite{cho2020x}.
However, this creates a mismatch between training and testing.
Results show that dense feature performs much worse than our two-level feature in text-to-image generation, despite its good performance in image-to-text generation.
Contrarily, using only the discrete feature performs much better than dense feature in text-to-image generation, however, not in image-to-text generation. 
Our design of two-level granularity feature inherits the advantages of dense features on image-to-text generation, and discrete features on text-to-image generation, and thus performs well on both tasks.

\noindent \textbf{What is the impact of the sequence-level training?}
We show that removing the sequence-level training and leaving a single stage of token-level training hurts performance significantly on both tasks.
For text-to-image generation, the CIDEr-D score drops from 122.6\%  to 107.9\%; for image-to-text generation, the CLIPScore drops from 77.2 to 73.4.
The results illustrate the effectiveness of sequence-level training to mitigate the ``exposure bias'' mismatch between training and testing.

\noindent \textbf{Do we need the CLIP-based loss?}
To verify the effect of our proposed CLIP-based loss for text-to-image generation in the image-level training, we replace the CLIP-based loss with a mean squared error (squared L2 norm) loss based on grid feature similarity.
From the results, we see that it hurts performance significantly on all text-to-image evaluation metrics.
The results highlight the effectiveness of CLIP-based loss to improve the semantic consistency between generated images and source text.
The performance of image-to-text generation is not influenced, demonstrating the robustness of our model.

\subsection{Quantitative Results}
We compare our approach with typical published uni-directional approaches with task-specific designs, and bi-directional approaches with or without task-specific designs as followed.
We choose models that are mainly trained with MS-COCO dataset as ours.
\begin{itemize}
    \item \textbf{DM-GAN}~\cite{zhu2019dm} is a typical text-to-image generation model with a GAN-based framework. It progressively generates images from low-resolution to high-resolution synthesis, and adaptively refines images with a memory module.
    \item \textbf{BUTD}~\cite{anderson2018bottom} is a image-to-text generation model with an encoder-decoder architecture based on CNN-LSTM networks. It uses a combined bottom-up and top-down attention mechanism to enable attention on objects.
    \item \textbf{Turbo-RL}~\cite{huang2018turbo} is a bi-directional image and text generation model. It is jointly trained with an LSTM network and a GAN network to boost both generators by enforcing cycle consistency in turbo learning.
    Since the authors did not report text-to-image generation results relevant to our metric, and we cannot find released codes, their results are not reported.
    \item \textbf{X-LXMERT}~\cite{cho2020x} is a vision-and-language pre-trained model for text-to-image generation, visual question answering, and visual reasoning tasks. X-LXMERT remains image captioning capability to sample sentences by masking and predicting word tokens. As the authors did not report their image-to-text results, we generate captions with a prefix word ``A'' as they suggested by their released code and model.
    \item \textbf{X-LXMERT-FT} is a model separately fine-tuned on image-to-text or text-to-image generation task from the released pre-trained X-LXMERT model with standard cross-entropy loss training. This provides a direct comparison with our model on feature representation and training strategy.
\end{itemize}

Table~\ref{tab:sota} provides comprehensive results.
Across two tasks, our model achieves significant performance gains over comparing methods in all metrics.
For task-specific uni-directional models, our model outperforms the typical image-to-text model \textbf{BUTD} by 2.5\% CIDEr-D score, despite being trained with much fewer images.
Our model also outperforms typical text-to-image model \textbf{DM-GAN} by 4.9 and 6.0 on CLIPScore and FID respectively.
The task-specific bi-directional model \textbf{Turbo-RL} achieves 74.8\% CIDEr-D score on MSCOCO Karpathy test set, which is far from satisfactory.
The inferior results achieved by task-specific works may indicate the limitation of RNN caption decoder or CNN image generator.

For Transformer-based approaches, the pre-trained \textbf{X-LXMERT} model achieves 41.0\% in CIDEr-D score and 15.2\% in BLEU@4, indicating the original model is not able to generate very accurate and fluent captions for MS-COCO dataset.
This large gap may be due to the mismatch between X-LXMERT's pre-training dataset (MS-COCO~\cite{lin2014microsoft}, VG~\cite{krishna2017visual} and VQA~\cite{antol2015vqa}) and downstream dataset (MS-COCO), so we fine-tune X-LXMERT with the standard cross-entropy loss training denoted by X-LXMERT-FT for fair comparisons.
X-LXMERT-FT do obtain large gains from 41.0\% to 100.9\% in CIDEr-D metric for image-to-text generation task, and from 37.0 to 33.9 in FID score for text-to-image generation task.
However, there is still a gap between X-LXMERT-FT and the task-specific models, and there is little gain over semantic consistency-aware metrics CLIPScore and R-precision.
Our model mitigates this gap by significantly improving CIDEr-D score from 100.9\% to 122.6\%.
Moreover, our model surpasses \textbf{X-LXMERT-FT} by 3.9 points in CLIPScore, 4.6/8.5 points in VilBERT-based R-precision, and 5.6/7.6 points in CLIP-based R-precision.
The results confirm the superiority of our model, and validates that a unified Transformer could result in similar breakthroughs in both image-to-text and text-to-image generations.

\begin{figure}[t]
    \centering
    \includegraphics[width=0.95\linewidth]{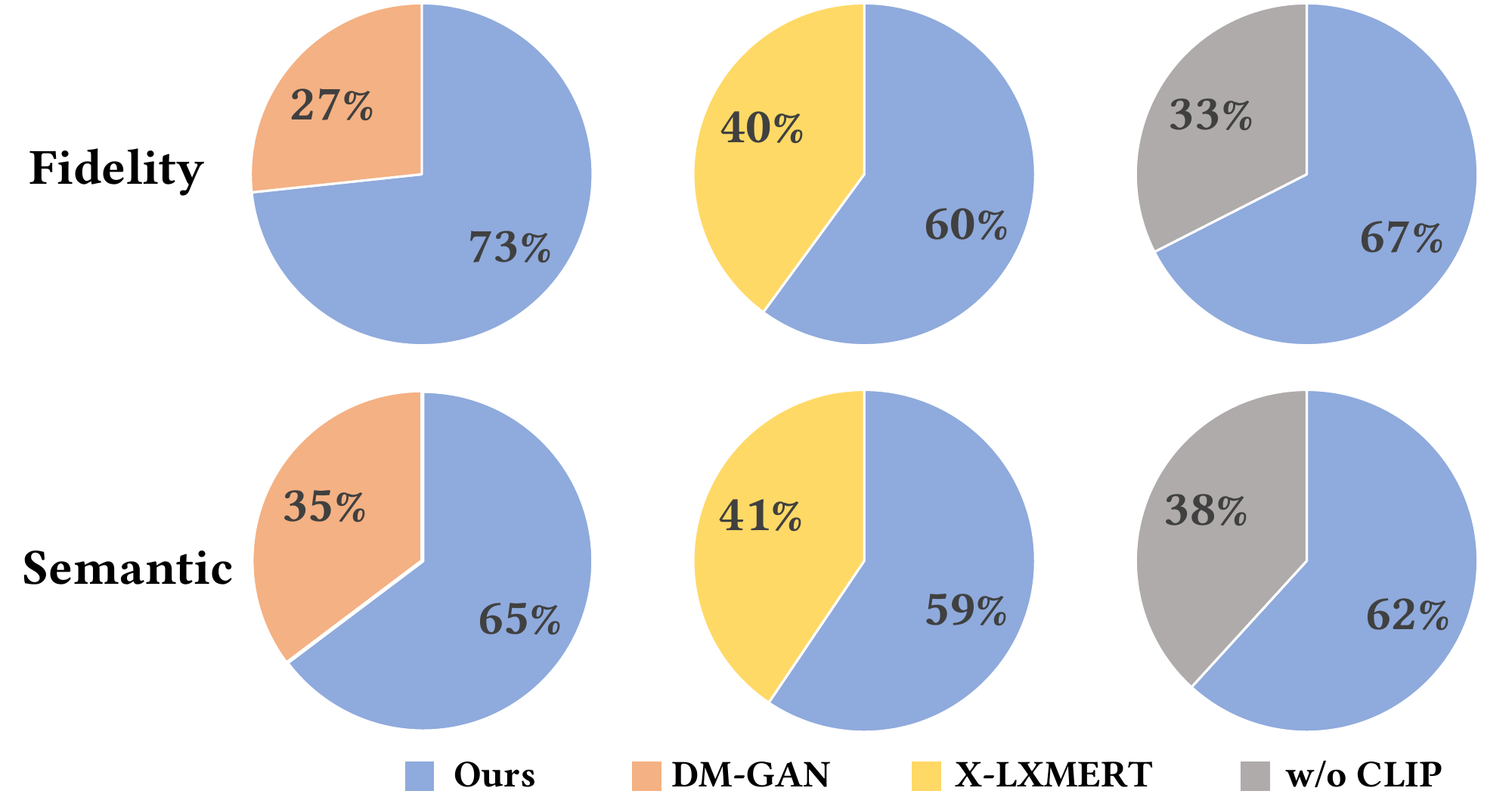}
    \caption{Human evaluation on text-to-image generation between our model, DM-GAN~\cite{zhu2019dm}, X-LXMERT~\cite{cho2020x} and our ablated model trained without CLIP-based loss (denoted by ``w/o CLIP''). 
    Our model clearly outperforms others in both aspects of fidelity and semantic of images.}
    \label{fig:user}
\end{figure}

\subsection{Human Evaluation}
To better evaluate the quality of generated images, we conduct a human evaluation to compare our method with existing works and visualize results in Figure \ref{fig:user}.
Specifically, we choose DMGAN~\cite{zhu2019dm} and X-LXMERT~\cite{cho2020x} for comparisons, which are the best-performing GAN-based and Transformer-based published works with released models, respectively.
We also compare with our ablated model without CLIP-based loss (denoted by ``w/o CLIP''). 
We randomly sample 300 captions from MSCOCO test set and generate images from each caption by different models.
During the evaluation, we provide a caption and an image pair generated by our model and other models in random order.
We invite ten volunteers who are proficient in English with over ten years of learning experiences and ask to select the one that (1) shows higher fidelity and (2) better matches the source caption semantically. 
As we can see from Figure \ref{fig:user}, our model significantly surpasses DM-GAN, X-LXMERT and ``w/o CLIP'' by about 37\%, 19\% and 29\% on an average of fidelity and semantic scores.
The results validate the superiority of our model.

\begin{figure*}[!hp]
    \centering
    \includegraphics[width=\linewidth]{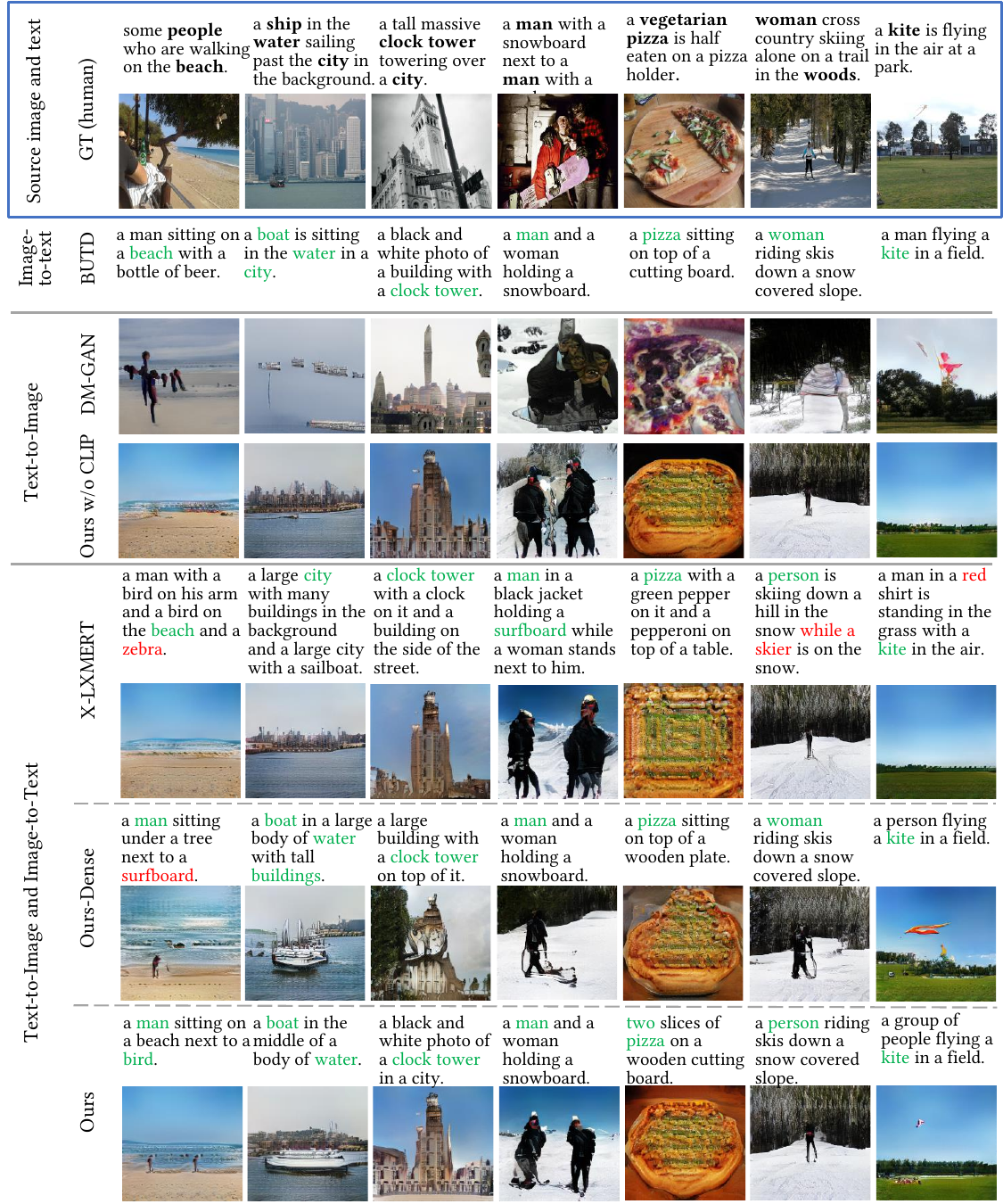}
    \caption{Examples of text-to-image and image-to-text generation results generated by different models from human written captions or real images (human ground-truths in \textcolor{blue}{blue} box). \textcolor{green}{Right} and \textcolor{red}{wrong} expressions in captions are highlighted in \textcolor{green}{green} and \textcolor{red}{red} respectively. Our model generates the most accurate and real images and text when compared with others. } \label{fig:example}
\end{figure*}

\subsection{Qualitative Examples}
Visual inspection of generated images and captions in Figure \ref{fig:example} convincingly shows the large quality improvement. 
For text generation, our model performs comparable with BUTD, which is  uni-directional approach. Compared with X-LXMERT, our model generates more accurate captions with better rationality. Some made-up objects like ``zebra'' and ``red shirt'' do not appear in our generated captions.
For image generation, our model generates images with more realistic objects and higher consistency with texts. Our model outperforms DM-GAN by a large margin on fidelity. Compared with X-LXMERT and w/o CLIP, our model learns better alignment with texts from CLIP-based loss (notice people on the beach, ship in the water, kite in the air.). Compared with the discrete-feature-based model, our model generates more realistic and smooth images.

\section{Conclusion}
In this work, we propose a unified multimodal Transformer for bi-directional image-and-text generation tasks.
Our proposal alleviates the expensive design efforts of task-specific models, and optimizes storage utilization compared to the design of two separate models for bi-directional tasks. 
To tackle the challenges of Transformer-based image-and-text generative models, we design a two-level granularity feature representation and a sequence-level training strategy.
The two-level granularity feature representation addresses the information loss issue caused by feature discretization process.
The sequence-level training strategy address the error accumulation in test time caused by the cross-entropy training.
Sufficient qualitative and quantitative experiments have shown the effectiveness of our approach.
With these benefits, our model could facilitate industrial applications for multimodal interactions with desirable performance.

\section{Acknowledgement}
This work was supported by the NSFC (U1811461), the Guangdong Natural Science Foundation (2018B030312002), and the Program for Guangdong Introducing Innovative and Entrepreneurial Teams under Grant NO.2016ZT06D211.

We would like to acknowledge Yiheng Xu and Zhicheng Huang for the helpful discussions.

\bibliographystyle{ACM-Reference-Format}
\balance
\bibliography{base}

\end{document}